\documentclass[sigconf,screen,authorversion,nonacm]{acmart}
\makeatletter
\if@ACM@anonymous
  \newcommand{\ifanonymize}[2]{#1}
\else
  \newcommand{\ifanonymize}[2]{#2}
\fi
\makeatother
\AtBeginDocument{%
  }

\setcopyright{cc}
\setcctype{by}
\copyrightyear{2026}
\acmYear{2026}
\acmDOI{XXXXXXX.XXXXXXX}
\acmConference[ACM e-Energy 2026]{17th ACM International Conference on Future and Sustainable Energy Systems}{June 22--25, 2026}{Banff, Canada}
\acmISBN{978-1-4503-XXXX-X/2026/06}

\usepackage{blindtext}
\usepackage{caption}
\usepackage{subcaption} 
\usepackage{siunitx}
\usepackage{booktabs}
\usepackage{bm}
\usepackage{enumitem}
\setlist[itemize]{align=parleft,left=3pt}

\begin{document}

\title[Explainable Load Forecasting with Covariate-Informed Time Series Foundation Models]{Explainable Load Forecasting with{\\}{ }Covariate-Informed Time Series Foundation Models}

\author{Matthias Hertel}
\authornote{Matthias Hertel and Alexandra Nikoltchovska contributed equally.}
\email{matthias.hertel@kit.edu}
\orcid{0000-0002-0814-766X}
\affiliation{%
  \institution{Karlsruhe Institute of Technology}
  \city{Karlsruhe}
  \country{Germany}
}

\author{Alexandra Nikoltchovska}
\authornotemark[1]
\email{alexandra.nikoltchovska@kit.edu}
\orcid{0009-0004-0293-6571}
\affiliation{%
  \institution{Karlsruhe Institute of Technology}
  \city{Karlsruhe}
  \country{Germany}
}

\author{Sebastian Pütz}
\email{sebastian.puetz@kit.edu}
\orcid{0009-0009-8468-4166}
\affiliation{%
  \institution{Karlsruhe Institute of Technology}
  \city{Karlsruhe}
  \country{Germany}
}

\author{Benjamin Schäfer}
\email{benjamin.schaefer@kit.edu}
\orcid{0000-0003-1607-9748}
\affiliation{%
  \institution{Karlsruhe Institute of Technology}
  \city{Karlsruhe}
  \country{Germany}
}

\author{Ralf Mikut}
\email{ralf.mikut@kit.edu}
\orcid{0000-0001-9100-5496}
\affiliation{%
  \institution{Karlsruhe Institute of Technology}
  \city{Karlsruhe}
  \country{Germany}
}

\author{Veit Hagenmeyer}
\email{veit.hagenmeyer@kit.edu}
\orcid{0000-0002-3572-9083}
\affiliation{%
  \institution{Karlsruhe Institute of Technology}
  \city{Karlsruhe}
  \country{Germany}
}

\renewcommand{\shortauthors}{M. Hertel, A. Nikoltchovska, et al.}

\begin{abstract}
Time Series Foundation Models (TSFMs) have recently emerged as general-purpose forecasting models and show considerable potential for applications in energy systems.
However, applications in critical infrastructure like power grids require transparency to ensure trust and reliability and cannot rely on pure black-box models.
To enhance the transparency of TSFMs, we propose an efficient algorithm for computing Shapley Additive Explanations (SHAP) tailored to these models.
The proposed approach leverages the flexibility of TSFMs with respect to input context length and provided covariates.
This property enables efficient temporal and covariate masking (selectively withholding inputs), allowing for a scalable explanation of model predictions using SHAP.
We evaluate two TSFMs -- Chronos-2 and TabPFN-TS -- on a day-ahead load forecasting task for a transmission system operator (TSO).
In a zero-shot setting, both models achieve predictive performance competitive with a Transformer model trained specifically on multiple years of TSO data.
The explanations obtained through our proposed approach align with established domain knowledge, particularly as the TSFMs appropriately use weather and calendar information for load prediction. Overall, we demonstrate that TSFMs can serve as transparent and reliable tools for operational energy forecasting.
\end{abstract}


\begin{teaserfigure}
  \includegraphics[width=\textwidth]{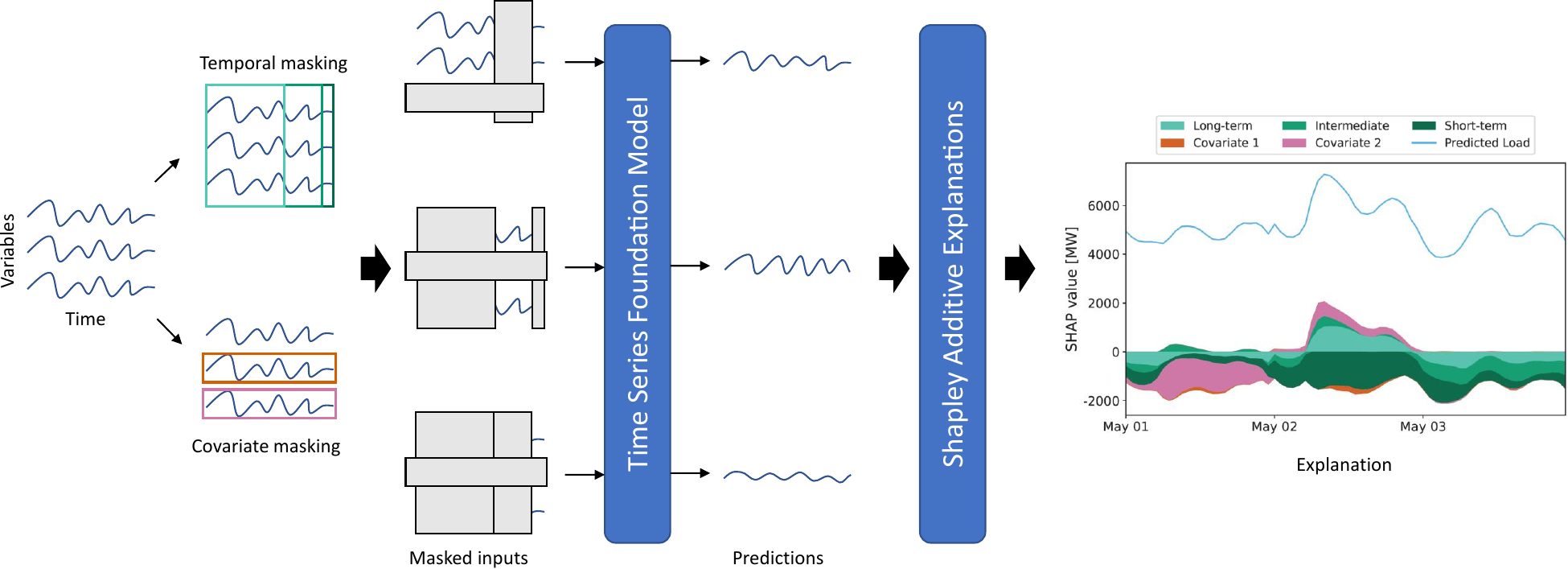}
  \caption{Overview of our approach to efficiently compute exact Shapley Additive Explanations (SHAP) for covariate-informed Time Series Foundation Models (TSFMs). We begin with historical hourly load observations accompanied by weather and calendar covariates. Our masking strategy operates along temporal and covariate dimensions — selectively removing historical time steps or entire covariates to create coalition samples. Each masked sample is then processed independently through Chronos-2 and TabPFN-TS to generate forecasts. Finally, SHAP values are computed by comparing predictions across coalitions, attributing the forecast to specific periods in the context and/or individual covariates.}
  \Description{Overview of our approach to efficiently compute exact Shapley Additive Explanations (SHAP) for covariate-informed Time Series Foundation Models (TSFMs). We begin with historical hourly load observations accompanied by weather and calendar covariates. Our masking strategy operates along temporal and covariate dimensions — selectively removing historical time steps or entire covariates to create coalition samples. Each masked sample is then processed independently through Chronos-2 and TabPFN-TS to generate forecasts. Finally, SHAP values are computed by comparing predictions across coalitions, attributing the forecast to specific periods in the context and/or individual covariates.}
  \label{fig:teaser}
\end{teaserfigure}

\received{29 January 2026}

\maketitle

\section{Introduction}

Accurate forecasting plays a central role in modern energy systems, as it is essential for balancing supply and demand, maintaining grid stability, and anticipating and mitigating grid congestions~\cite{hong_energy_2014, haben_review_2021}. 
Reliable forecasts of electrical load, renewable generation, and energy prices enable a wide range of operational and planning applications, including the scheduling of redispatch measures, demand-side management, storage dispatch, and the operation of energy management systems. 
In recent years, forecasting performance has been substantially improved through the use of deep learning models. 
While these models achieve good prediction accuracy, they lack transparency, which limits their acceptance in safety-critical and operational settings ~\cite{machlev_explainable_2022}.
This challenge has become increasingly urgent with the introduction of regulatory frameworks such as the EU AI Act and guidelines for AI deployment in critical infrastructure, which emphasize the need for transparency and accountability in high-stakes applications~\cite{european_commission_artificial_2024, dhs_ai_framework_2024}.

Time Series Foundation Models (TSFMs)~\cite{kottapalli_hubli_2025} have recently emerged as a new class of large-scale deep learning models, built predominantly on Transformer architectures and pretrained on large and diverse collections of time series data. Unlike conventional deep learning approaches for forecasting, which are typically trained from scratch on task-specific datasets, TSFMs leverage large-scale pretraining to achieve strong zero-shot and few-shot generalization across tasks and domains. More recent architectures, such as Chronos-2~\cite{ansari_chronos-2_2025} and TabPFN-TS~\cite{hoo_tables_2025}, extend this paradigm by supporting the integration of covariates — additional time-varying features that influence the target (sometimes also referred to as exogenous features or auxiliary information) — which is particularly relevant for energy system applications.

Like previous work~\cite{ansari_chronos-2_2025, hoo_tables_2025}, we define models as \emph{univariate} when they forecast a single time series, whereas \emph{multivariate} models forecast multiple time series together.
Models that incorporate covariates in addition to past values from the target time series are called \emph{covariate-informed}.

Despite their promising performance, covariate-informed TSFMs remain largely opaque, and their increasing complexity exacerbates the challenge of understanding and trusting their predictions.
Developing and applying suitable Explainable AI (XAI) methods is therefore a key requirement for the deployment of TSFMs in operational energy system contexts. 
Post-hoc explainability methods such as Shapley Additive Explanations (SHAP) ~\cite{lundberg_unified_2017} offer theoretically grounded insights into model behavior, but their computational cost poses a major barrier when applied to large-scale time series models with long input horizons and multiple covariates.

To address this gap, we introduce an efficient SHAP-based explainability algorithm tailored to TSFMs and demonstrate its applicability to load forecasting, a key challenge in the energy domain~\cite{khan_mahmood_2016}. 
We apply our approach to two architecturally distinct TSFMs -- Chronos-2~\cite{ansari_chronos-2_2025} and TabPFN-TS~\cite{hoo_tables_2025} -- using operational electrical load data from \ifanonymize{a major TSO}{Baden-Württemberg, Germany}.
The overall procedure of our approach is illustrated in Figure~\ref{fig:teaser}.
We make the following key contributions:

\begin{enumerate}
    \item We compare two covariate-informed TSFMs -- Chronos-2 and TabPFN-TS -- for electrical load forecasting on recent operational data from a specific TSO region, and benchmark their performance against state-of-the-art models trained from scratch.
    \item We propose an efficient SHAP algorithm for TSFMs, using temporal and covariate masking to enable the estimation of SHAP values without the need for sampling from background data.
    \item We empirically demonstrate that Chronos-2 and TabPFN-TS make meaningful use of covariates, such as calendar effects and weather variables, and that the resulting explanations are consistent with established domain knowledge.
\end{enumerate}

The remainder of this paper is structured as follows: Section~\ref{sec:related-work} reviews related work on load forecasting, TSFMs, and post-hoc explainability methods. 
The proposed approach for efficiently computing SHAP-based explanations is introduced in Section~\ref{sec:methodology}, followed by a description of the experimental setup in Section~\ref{sec:experiments}, including the dataset, preprocessing steps, TSFM configurations and evaluation metrics. 
Section~\ref{sec:performance} assesses the performance of TSFMs on recent operational load data.
Section~\ref{sec:explanations} presents the explainability analysis, examining TSFM behavior through global feature importance patterns and local explanations of individual predictions. 
Finally, Section~\ref{sec:conclusion} draws conclusions and discusses future research directions.

\section{Related work}
\label{sec:related-work}

\paragraph{Machine Learning for Load Forecasting} 
While traditional load forecasting methods relied primarily on statistical approaches ~\cite{hong_energy_2014}, in recent years, there has been a shift towards more complex deep learning architectures ~\cite{hong_energy_2020, scheidt_data_2020, haben_review_2021, giacomazzi_short-term_2023}.
Recent work showed that global models -- trained on multiple time series, such as load profiles from multiple buildings or substations -- can capture shared patterns across different contexts and achieve superior generalization compared to local models trained on individual time series \ifanonymize{\cite{montero-manso_principles_2021, grabner_global_2023}}{~\cite{montero-manso_principles_2021, grabner_global_2023, hertel_transformer_2023, hertel_comparison_2025}}.
In addition, the effectiveness of load forecasting models improves when covariates are incorporated alongside historical load observations.
Variables such as temperature, solar irradiation, and calendar information have been shown to improve prediction accuracy by capturing the external factors that drive consumption patterns ~\cite{haben_review_2021}.
It is particularly important to also include covariates for future time steps, which can be either known in advance, such as holidays and calendar information, or can be forecasts themselves, such as weather predictions.

\paragraph{Time Series Foundation Models (TSFMs)}
TSFMs take the global modeling paradigm further by pretraining on large and diverse datasets that often include synthetic data, enabling them to identify patterns across domains when applied to specific forecasting tasks ~\cite{liang_foundation_2024}.
Initial TSFM architectures focused primarily on univariate forecasting, learning temporal patterns from historical values of the target series alone. 
Chronos ~\cite{ansari_chronos_2024} introduced the application of language model tokenization to time series, treating forecasting as a sequence-to-sequence task based on quantized observations.
Lag-Llama ~\cite{rasul_lag-llama_2023} employs a decoder-only transformer architecture pretrained on large-scale univariate time series.
TimesFM ~\cite{das_decoder-only_2024} uses a similar decoder-only design, emphasizing patching mechanisms to capture multi-scale temporal patterns.
However, as with load forecasting, many real-world forecasting tasks require the integration of covariates to achieve accurate predictions.
Recognizing this need, recent TSFM architectures have been extended to handle covariates.
Chronos-2 ~\cite{ansari_chronos-2_2025} extends the original Chronos framework to explicitly incorporate covariates during both pretraining and inference while maintaining strong zero-shot capabilities.
TabPFN-TS~\cite{hoo_tables_2025} takes a different approach by extending tabular foundation models to time series forecasting, using in-context learning to adapt to new tasks without gradient-based fine-tuning.
While Moirai~\cite{woo_liu_2024} also claims to be able to handle covariates, recent evaluations suggest it does not effectively leverage future covariates in load forecasting applications~\cite{kreusel_evaluating_2025}.

We select Chronos-2 and TabPFN-TS, representing state-of-the-art Transformer-based and tabular foundation model paradigms, respectively, as both have shown impressive results on established forecasting benchmarks, specifically on covariate-informed tasks~\cite{shchur_fev-bench_2025, aksu_gift-eval_2024}.
This comparison allows us to examine whether feature importance patterns are consistent across different modeling paradigms, with extension to further TSFMs left for future work.

Despite their capability to generalize to unseen data, TSFMs have primarily been developed and evaluated on standard benchmark datasets.
Recent work by ~\citet{meyer_kaltenpoth_2025} raises concerns about this paradigm, suggesting that TSFM performance can substantially drop on datasets that differ from those used in the original publications.
Initial applications of TSFMs to energy load forecasting have started filling this gap and have demonstrated their potential in this domain.
~\citet{meyer_benchmarking_2024} and ~\citet{lin_comparative_2025} showed competitive performance of univariate TSFMs on short-term load forecasting benchmarks, while ~\citet{kreusel_evaluating_2025} extended these evaluations to covariate-informed settings, highlighting the promise of covariate-informed TSFMs for energy applications.
Recent findings question whether TSFMs can deliver on their "one-size-fits-all" promise, showing that zero-shot performance depends heavily on the alignment between pretraining and target data, and that lightweight specialized models can outperform fine-tuned TSFMs under domain shift or concept drift~\cite{karaouli_coquenet_2025}.
Given that load forecasting involves region-specific weather dependencies and evolving consumption patterns, continued evaluation of covariate-informed TSFMs on recent operational data from specific TSO regions remains valuable.
Therefore, for our first contribution we evaluate Chronos-2 and TabPFN-TS on electrical load data from \ifanonymize{one specific European TSO}{the German TSO TransnetBW}.

\paragraph{Post-hoc Explainability for Time Series Models}
While the evaluation of TSFMs on domain-specific data addresses questions of predictive performance, their deployment in critical infrastructure, such as power grids, raises an equally important concern: model transparency. 
TSFMs, similar to deep learning methods, operate as black boxes whose learning and decision processes remain hidden from energy domain experts such as system operators. 
Operational energy forecasting requires not only accurate forecasts, but also validation that models appropriately respond to known physical relationships, such as temperature-load dependence and calendar-driven consumption patterns.
Post-hoc explainability techniques have therefore become essential tools for interpreting complex forecasting models without sacrificing their predictive power, enabling domain experts to inspect model behavior and build the operational trust necessary for deployment~\cite{machlev_explainable_2022, baur_explainability_2024}.

Among post-hoc methods, SHAP ~\cite{lundberg_unified_2017} has been widely adopted due to its theoretical foundations in cooperative game theory and its model-agnostic property. SHAP is a unified framework for explaining model predictions, based on Shapley values from game theory ~\cite{shapley_lloyd_s_value_1953}, used to fairly distribute the total payoff of a cooperative game between players, when their individual contributions vary and they interact with one another. 
These values satisfy fairness axioms that make them particularly meaningful for post-hoc interpretability. For example, efficiency ensures that attributions sum to the total prediction shift, and symmetry ensures that features with identical contributions receive identical attributions.
In machine learning, features act as players, with the payoff being the difference between prediction and the base value (expected value or the average prediction of the model across the dataset).


Despite these desirable theoretical properties, SHAP is computationally expensive in practice. The method requires many model evaluations based on feature subsets, called coalitions. As most models cannot be evaluated with subsets of the features provided during training, a common approach is to sample absent features multiple times from a background dataset. This results in the need for many model evaluations, which is especially costly for large models such as TSFMs.

To address these computational limitations, several methods improve the efficiency of SHAP for time series. 
TimeSHAP ~\cite{bento_timeshap_2021} groups consecutive time steps to reduce the number of coalitions, while WindowSHAP ~\cite{nayebi_windowshap_2023} uses sliding windows for temporal localization. 
ShapTime~\cite{zhang_shaptime_2024} leverages temporal structure for acceleration, and SHAPformer~\cite{hertel_putz_2025} combines Transformers with temporal grouping. 
Further speedups are achieved by avoiding sampling, either replacing absent features with a base value ~\cite{bento_timeshap_2021} or restricting the models access to absent values with attention manipulation ~\cite{hertel_putz_2025, deiseroth_atman_2023}.
However, these methods either still rely on background sampling or are tied to specific architectures, making them computationally infeasible for general-purpose TSFMs. Our work adopts the temporal grouping idea from TimeSHAP and WindowSHAP, while eliminating background sampling by exploiting the flexibility of TSFMs to handle arbitrary input contexts.

Most TSFMs are based on the Transformer architecture~\cite{vaswani_attention_2017}, which allows to visualize the attention weights to highlight important inputs as an explanation~\cite{hertel_evaluation_2022}. 
However, attention is only a proxy for feature importance~\cite{jain_attention_2019,bibal_is_2022} and highlights inputs without revealing their directional influence on predictions. 
Therefore, we use SHAP, which provides directional, quantitative feature attributions.

The intersection of explainability and TSFMs remains largely underexplored. ~\citet{pandey_internal_2025} examined internal representations of TSFMs through synthetic scenarios, but focused on model behavior rather than real-world applications. 
~\citet{boileau_helluy_2025} developed inherently interpretable TSFM architectures, which represent a different design philosophy than post-hoc explanation of existing high-performance models. 
~\citet{longo_interpretable_2024} introduced a method for explaining TabPFN predictions, but this work focuses on static tabular tasks and does not address the temporal dimension inherent in time series forecasting. 

Building on these foundations, we address the need for efficient XAI methods tailored to covariate-informed TSFMs, where both temporal dependencies and the effects of covariates must be considered. 
This motivates our second contribution: leveraging the internal structure of TSFMs for an efficient SHAP adaptation using temporal and covariate masking that eliminates the need for background data sampling. 
Our third contribution empirically demonstrates that covariate-informed TSFMs meaningfully utilize covariates like calendar effects and weather variables, with explanations that are consistent with established domain knowledge.

\section{Methodology}
\label{sec:methodology}

We develop efficient SHAP-based explanations by leveraging the internal structures of Chronos-2 and TabPFN-TS to implement architecture-specific masking strategies, eliminating the need for background data sampling. The methodology is depicted in Figure~\ref{fig:teaser}.
To establish the foundation for our approach, we first briefly revisit the fundamentals of SHAP. 
Then, we describe how the TSFMs process input data, and finally outline our feature grouping and masking strategies.
Concise summaries of the model architectures are provided in Appendix~\ref{app:methods}, with full technical details in the original publications~\cite{ansari_chronos-2_2025, hoo_tables_2025}.

\paragraph{SHAP}


For a model $f$ trained on a feature set $V = \{v_1, v_2, \ldots, v_n\}$ consisting of $n$ features, the SHAP value for feature $v_i$ is calculated as:
\begin{equation}
    \textit{SHAP}(v_i) = \sum_{S \subseteq V \setminus \{v_i\}} \dfrac{(n - 1 - |S|)! \cdot |S|!}{n!} \cdot (f(S \cup \{v_i\}) - f(S)),
\end{equation}
where $S$ represents all possible subsets (coalitions) of features excluding $v_i$, and $f(S \cup \{v_i\}) - f(S)$ quantifies the marginal contribution of $v_i$ to coalition $S$.

Computing exact SHAP values requires evaluating all $2^n$ possible feature coalitions, resulting in exponential computational complexity of $O(2^n)$. 
Since this becomes intractable even for a moderate number of features, in practice, SHAP values are typically estimated using sampling-based approximations. However, these approximations introduce variance and can require thousands of model evaluations to achieve stable estimates, making them computationally expensive for time series forecasting applications.

An additional challenge is that classical machine learning models require all features they have been trained on for inference. Evaluating $f(S)$ for a subset of features $S \subset V$ then requires handling the missing feature values. In practice, this is usually done by computing the marginal contribution of a feature by marginalizing over the missing features, either by sampling from a background dataset or by integrating over their empirical distribution.

\paragraph{TSFM inputs} In contrast to classical machine learning models, TSFMs are flexible in the context length and covariates provided to the model. This allows to evaluate them on subsets of the data without sampling absent features from background data.
A TSFM evaluated with context length $C$ and forecast horizon $H$ receives the past $C$ values of the target time series as input (denoted as $X \in \mathbb{R}^C$), together with an arbitrary number of covariates $Z_i$. These covariates can either be available for the past only ($Z_i \in \mathbb{R}^C$), or extend into the forecast horizon ($Z_i \in \mathbb{R}^{C+H}$) when they are known in advance, such as calendar features or weather forecasts.

\paragraph{Feature grouping} Computing exact SHAP values requires to evaluate the TSFM with $2^n$ feature subsets for $n$ features. As this is not feasible for large amounts of features, we instead group the features into $N$ subsets, with $N \ll n$, and evaluate the TSFM only $2^N$ times.
We incorporate two grouping mechanisms:
\emph{Temporal grouping} splits the past load time series into consecutive patches, in order to analyze the effect of the load in different past periods. In particular, we introduce one group for the last day (hours $-24$ to $-1$ relative to the time the prediction is made), short-term (hours $-168$ to $-25$), intermediate (hours $-672$ to $-169$) and long-term (hours $<-672$) periods.
\emph{Covariate grouping} builds one group for each covariate, which is used to analyze the effect of the covariate on the prediction.
In total, we analyze four temporal groups and $|Z|$ covariate groups.

\paragraph{Masking for Chronos-2}
To evaluate a given feature subset, the groups not belonging to it are \emph{masked}, i.e., withheld from the model input. 
\emph{Temporal masking} works in two ways. To mask the first $k$ values of the context window, the context length gets reduced to $C-k$, so that the masked values are not contained in the context.
To mask intermediate or the most recent values, the load values are set to not-a-number (NaN), which indicates missing values. Internally, Chronos-2 sets missing values to zero and uses a special token to distinguish between missing values and measured zeros.
\emph{Covariate masking} works by removing masked covariates entirely from the model input.
When all data is masked so that the model input is empty, a base prediction is made. For this, we use the mean load from the previous 48 weeks, which corresponds to the model's maximum input length.

\paragraph{Masking for TabPFN-TS} To mask the first $k$ values of the context window, \emph{temporal masking} for TabPFN-TS works the same way as for Chronos-2.
To mask intermediate or the most recent values, corresponding rows are dropped from the model input entirely, due to the inability of TabPFN-TS to handle missing values in the target variable.
\emph{Covariate masking} removes masked covariates entirely from the model input, equivalent to the Chronos-2 approach.
Similar to Chronos-2, we use the mean load from the previous year as a base prediction when all data is masked.


\section{Experimental Setup}
\label{sec:experiments}
\paragraph{Dataset}

The dataset comprises the electrical load of \ifanonymize{a European TSO}{the German TSO TransnetBW}, downloaded from the ENTSO-E Transparency Platform~\cite{entsoe_transparency_2015} for the period January 2015 -- September 2025, and converted into hourly resolution.
The dataset is enriched with weather data from the ERA5 reanalysis model~\cite{hersbach_bell_2020} \ifanonymize{for the geographic region corresponding to the TSO service area.}{for the DE1 region comprising Baden-Württemberg, which has a typical peak load of roughly \qty{14}{\giga\watt} ~\cite{klobasa_angerer_2014}.}
From the weather dataset, the solar irradiance and the outside temperature are used as covariates.
In practice, the reanalysis data is not available at prediction time, but it is a common approach to use weather data as perfect forecasts in the absence of historical weather forecasts ~\cite{haben_review_2021}.
A third covariate is used to encode holidays, set to 1 for Sundays and public holidays, and to 0 otherwise.
Sundays and holidays are grouped into a single indicator because both exhibit similar reduced-demand patterns, and treating them as separate categories would result in a sparse feature with too few occurrences for reliable generalization~\cite{ziel_2018}.

The last year of data (October 2024 -- September 2025) is used as test data for the experiments, the year before (October 2023 -- September 2024) as validation data and everything before (January 2015 -- September 2023) as training data to train models from scratch.

\paragraph{Forecasting task \& metrics}
To evaluate forecast performance, a forecast is made at every hour in the test set with a horizon of 24 hours.
We evaluate the models as point forecasters, taking the median of the predicted distribution as the forecast.
Our focus is on evaluating and explaining point predictions, as these are the primary outputs used in operational forecasting decisions. Extending the analysis to probabilistic forecasts is left for future work.
We use three different metrics to evaluate the forecast accuracy: mean absolute error (MAE), root-mean squared error (RMSE), and mean absolute percentage error (MAPE), see Appendix ~\ref{app:metrics} for more details.
To explain models, we generate explanations of the forecasts made at midnight, so that each hour in the test set is predicted exactly once and we do not get multiple explanations for the same hour. Note that the concatenation of the explained forecasts comprises the entire year without overlaps.

\paragraph{Chronos-2} We evaluate Chronos-2 with the default hyperparameters in a zero-shot setting. Chronos-2 is a probabilistic model returning quantile predictions, which we turn into point forecasts by taking the median of the predicted distribution. We test different context lengths ranging from one week to the model's maximum of 8192 hours, and find that longer context lengths improve the forecast accuracy (see Appendix~\ref{app:context-lengths} for the detailed results). We report forecasting results for a univariate Chronos-2 and for Chronos-2 with the three covariates described above, in order to quantify their contribution.

\paragraph{TabPFN-TS} We evaluate TabPFN-TS following the steps proposed in the original publication~\cite{hoo_tables_2025}, which include generating a running index, calendar features (day of week, month, hour encoded as cyclic sine/cosine pairs), and automatic seasonal features that identify domain-specific periodicities beyond standard calendar cycles.
We construct an ensemble of two models: one trained on normalized targets and one trained on power transformed targets (Box-Cox transformation ~\cite{box_cox_1964}). 
Similar to Chronos-2, we test different context lengths (see Appendix ~\ref{app:context-lengths}) and report results for both univariate TabPFN-TS and covariate-informed TabPFN-TS to quantify the improvement gained with covariates.

\paragraph{Non-TSFM models}
We compare the TSFMs to a simple baseline and to Transformer models trained from scratch on the TSO data. The baseline predicts the observed load from the last day of the same type, distinguishing between workday, Saturday and Sunday/holiday as the types.
\ifanonymize{}{The \emph{Transformer} architecture is inspired by ~\citet{hertel_evaluation_2022}.}
\ifanonymize{The \emph{Transformer}}{It} is an encoder-decoder model with two layers each and a hidden dimension of $128$ and four attention heads, trained from scratch with two weeks context length using the AdamW optimizer with a batch size of $128$, a learning rate of $0.0001$ and early stopping with ten epochs patience.
We report results for two Transformer variants with different amounts of training data, namely one year (October 2022 -- September 2023) and the full training data comprising $8.75$ years (January 2015 -- September 2023).

\section{Benchmarking TSFMs on Recent Operational Load Data}
\label{sec:performance}
In this section, we analyze the forecast accuracy and the model runtimes to evaluate the applicability of TSFMs to load forecasting.

\paragraph{Forecast accuracy}

Table~\ref{tab:results} reports the forecasting results on the test set. All models outperform the baseline across all error metrics. The Transformer trained on the full data achieves the lowest errors, while the same architecture trained on one year of data shows substantially higher errors. Both TSFMs -- TabPFN-TS and Chronos-2 -- achieve lower errors than the Transformer trained on one year of data.
Compared to the one-year Transformer, TabPFN-TS reduces the MAE by 23.5\,\% and Chronos-2 by 26.8\,\%.
When trained on the full data, the Transformer is better than the TSFMs, but only by a small margin.
Between the two TSFMs, Chronos-2 achieves slightly lower errors than TabPFN-TS across all metrics in the covariate-informed setting. For both TSFMs, incorporating covariates leads to better accuracy than the univariate setting, highlighting the importance of incorporating exogenous information.
The MAE improvement by using covariates is 27.0\,\% for Chronos-2 and 31.5\,\% for TabPFN-TS.
Note that all models benefit from ERA5 reanalysis weather inputs, which represent historically reconstructed rather than forecasted conditions, so reported results represent an upper bound on operational accuracy.

Both models benefit from longer contexts, and Chronos-2 outperforms TabPFN-TS for all tested context lengths.
Notably, TabPFN-TS needs at least two weeks of context to beat the baseline, whereas Chronos-2 outperforms the baseline already with one week of context.
Potentially, this is due to the fact that TabPFN is pretrained as a tabular model, so it needs more context to infer the daily and weekly pattern inherent in the load time series, whereas Chronos-2 is pretrained on time series with various seasonalities.
Results for the TSFMs with different context lengths are reported in Appendix~\ref{app:context-lengths}. 

Despite the strong overall forecasting performance of both models, they encounter difficulties on specific days in proximity to public holidays, including October 4, December 23, January 7, June 20, and June 23. 
This suggests potential for improvement through feature engineering, such as incorporating covariates for long weekends, school holidays, or other periods of reduced economic activity. 
For the complete year of forecasts made daily at midnight, see Figure~\ref{fig:monthly-shap-values-chronos} for Chronos-2 and Figure~\ref{fig:monthly-shap-values-tabpfn} for TabPFN-TS in Appendix~\ref{app:local-explanations}, each evaluated with the best model configuration.

Overall, the results highlight the importance of integrating covariates into TSFMs for accurate load forecasting.
TSFMs are especially useful in scenarios with little training data, as they outperform a Transformer trained from scratch on one year of data.
When lots of data is available, models trained from scratch can achieve better results than zero-shot TSFMs, but the performance difference is small and the TSFMs have the advantage that they are easy to use without requiring any training, which makes them appealing in practice.

\begin{table}[hbt]
    \centering
    \caption{Forecasting results on the test set, with the best result per metric highlighted in bold and second best in italics. The TSFMs are evaluated in both univariate (univ.) and covariate-informed settings.}
    \label{tab:results}
    \begin{tabular}{lcccc}
    \toprule
    Model & Training & MAE  & RMSE & MAPE \\
          & data     & [\si{\mega\watt}] & [\si{\mega\watt}] & [\%] \\
    \midrule
    Baseline    & {---} &  325.9 & 451.2 & 5.24 \\
    \midrule
    Transformer & 1 year & 213.4 & 318.9 & 3.32 \\
    Transformer & 8.75 years & \textbf{150.2} & \textbf{206.8} & \textbf{2.34} \\
    \midrule
    TabPFN-TS (univ.) & {---} & 238.1 & 398.8 & 3.74 \\
    TabPFN-TS & {---} & 163.2 & 250.5 & 2.55 \\
    \midrule
    Chronos-2 (univ.) & {---} & 214.0 & 356.9 & 3.40 \\
    Chronos-2   & {---} & \textit{156.2} & \textit{229.1} & \textit{2.44} \\
    \bottomrule
    \end{tabular}\\
\end{table}

\paragraph{Runtimes}

Table~\ref{tab:runtimes} reports the runtimes for model training, inference and explanation.
Training the Transformer from scratch on the full training dataset takes less than 20 minutes. This computational cost is not needed with TSFMs, which are pretrained on large diverse datasets and applicable zero-shot to new data without the need for further training.
Chronos-2 has more parameters than the Transformer trained from scratch, and therefore needs more time to generate a prediction.
However, both models are very efficient and allow to predict in a few milliseconds.
TabPFN-TS, on the other hand, is much slower, needing multiple seconds to create a prediction.
To generate an explanation, the TSFMs are run with all $2^7=128$ combinations of the seven temporal and covariate masks.
Explaining a forecast takes less than $128\times$ the prediction time, because masking reduces the context length and the number of covariates in the model input.

\begin{table}[htb]
    \centering
    \caption{Training runtime, runtime to generate a single prediction, and runtime to generate a single explanation, with the best result highlighted in bold.}
    \label{tab:runtimes}
    \begin{tabular}{lSSSS}
    \toprule
    Model & {Training} & {Training} & {Prediction} & {Explanation} \\
          & {data}     & {[min]}    & {[ms]}        & {[s]}         \\
    \midrule
    Transformer & {8.75 years} & 19.2 & \textbf{6.8} &  {---} \\
    TabPFN-TS   & {---} &  {---} &  5521.2 & 366.84 \\
    Chronos-2   & {---} &  {---} & 65.6 & \hspace{0.17cm} \textbf{5.12} \\
    \bottomrule
    \end{tabular}
\end{table}


\section{Explaining TSFMs for Load Forecasting}
\label{sec:explanations}

Having established that TSFMs show competitive performance, we next investigate the SHAP-based global and local explanations of the two TSFMs to understand how the provided data influences the predictions.

\subsection{Global explanations}

\paragraph{Feature importance} Table~\ref{tab:feature-importance} presents the feature importance for the two TSFMs, calculated as the percentage of the absolute SHAP values over the test set.
The past load is by far the most important feature, with 89\,\% total importance for Chronos-2 and 87\,\% for TabPFN-TS.
For Chronos-2, the four temporal windows have similar importance, despite their different lengths. This indicates that recent information is important for the prediction, as the last day alone is almost as important as the whole long-term context.
TabPFN-TS also emphasizes the last day, but gives more relevance to the long-term context.
Potentially, this can be explained by the fact that TabPFN-TS is built on TabPFN, which is not a time series model per se, and therefore requires more context to infer time series patterns.
Regarding the covariates, the two TSFMs agree on the order of importance, with holiday being the most important, followed by temperature, and irradiance being the least important.
TabPFN-TS gives more importance to the holiday covariate than Chronos-2, whereas the importance of the other two covariates is similar for both TSFMs.

\begin{table}[htb]
    \centering
    \caption{Feature importance on the test set.}
    \label{tab:feature-importance}
    \begin{tabular}{lSSS}
    \toprule
    {Feature Group} & {Chronos-2} & {TabPFN-TS} \\
    \midrule
    \textbf{Past load} \\
    ~~Long-term (weeks $\leq$ -5)   & 23.64\,\% & 29.15\,\% \\
    ~~Intermediate (weeks -4 to -2) & 23.39\,\% & 18.67\,\% \\
    ~~Short-term (days -7 to -2)    & 19.88\,\% & 18.63\,\% \\
    ~~ Last day (day -1)        & 22.30\,\% & 20.60\,\% \\
    \midrule
    \textbf{Covariates} \\
    ~~Holiday                       & 4.51\,\%  & 6.46\,\%  \\
    ~~Temperature                   & 3.55\,\%  & 3.79\,\%  \\
    ~~Irradiance                    & 2.74\,\%  & 2.70\,\%  \\
    \bottomrule
    \end{tabular}
\end{table}

\paragraph{Covariate dependence}

In order to analyze how the TSFMs use covariates, Figure~\ref{fig:feature-dependence} shows the SHAP values of Chronos-2 and TabPFN-TS in dependence of the three covariates. The x-axis shows the feature values at the predicted hour (or the difference to the value 24 hours before) and the y-axis shows the corresponding SHAP values. A positive SHAP value indicates that the feature increased the prediction, and a negative SHAP value indicates a decreased prediction. The dot color is used to show an interacting variable.

\begin{figure*}[ht!]
    \centering
    \begin{subfigure}{0.49\textwidth}
        \caption{Chronos-2}
        \vspace{-0.1cm}
        \includegraphics[width=\textwidth]{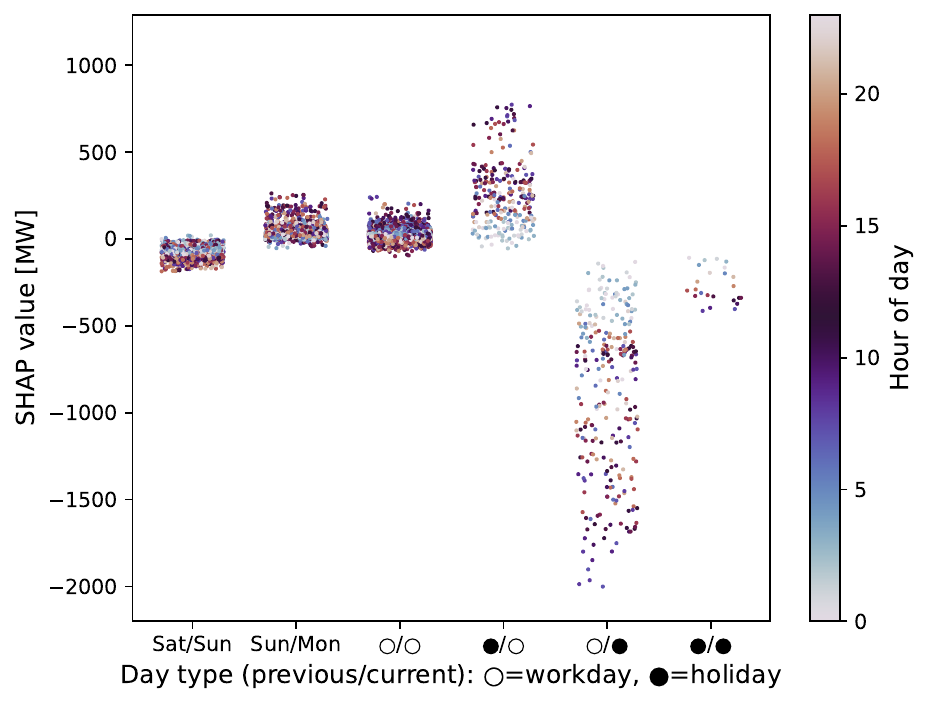}
        \label{fig:holiday-chronos2}
    \end{subfigure}
    \begin{subfigure}{0.49\textwidth}
        \caption{TabPFN-TS}
        \vspace{-0.1cm}
        \includegraphics[width=\textwidth]{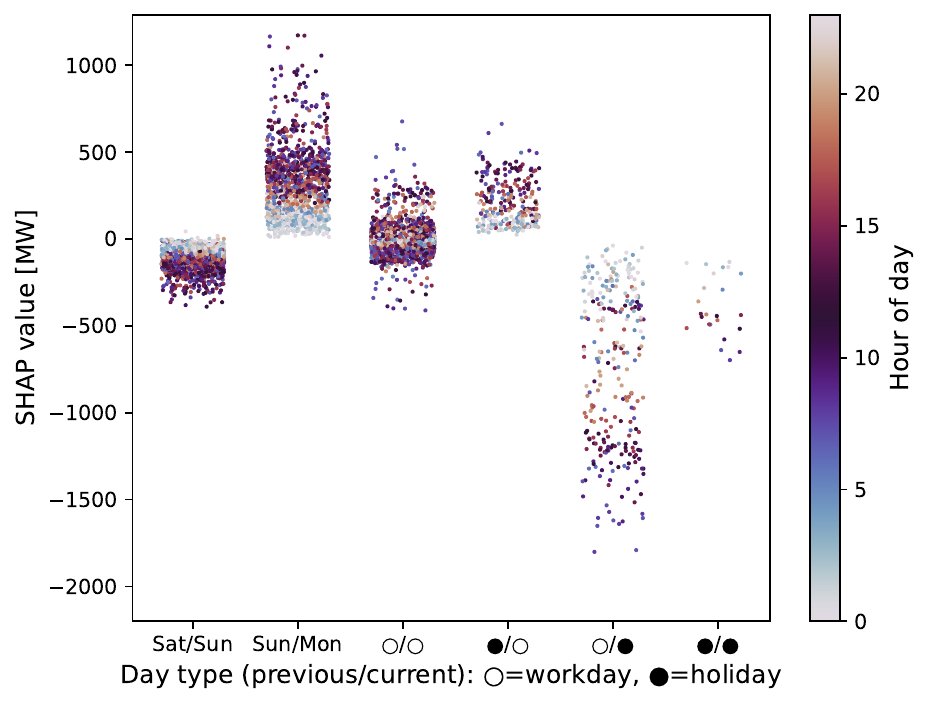}
        \label{fig:holiday-tabpfn}
    \end{subfigure}\\ \vspace{-0.5cm}
    \begin{subfigure}{0.49\textwidth}
        \includegraphics[width=\textwidth]{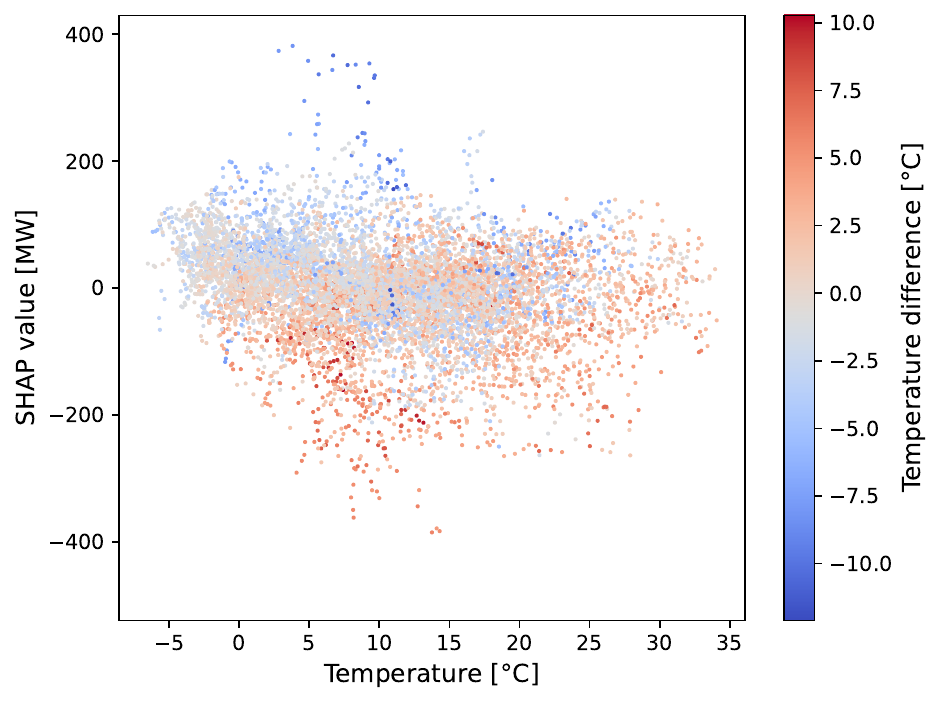}
        \label{fig:temperature-chronos2}
    \end{subfigure}
    \begin{subfigure}{0.49\textwidth}
        \includegraphics[width=\textwidth]{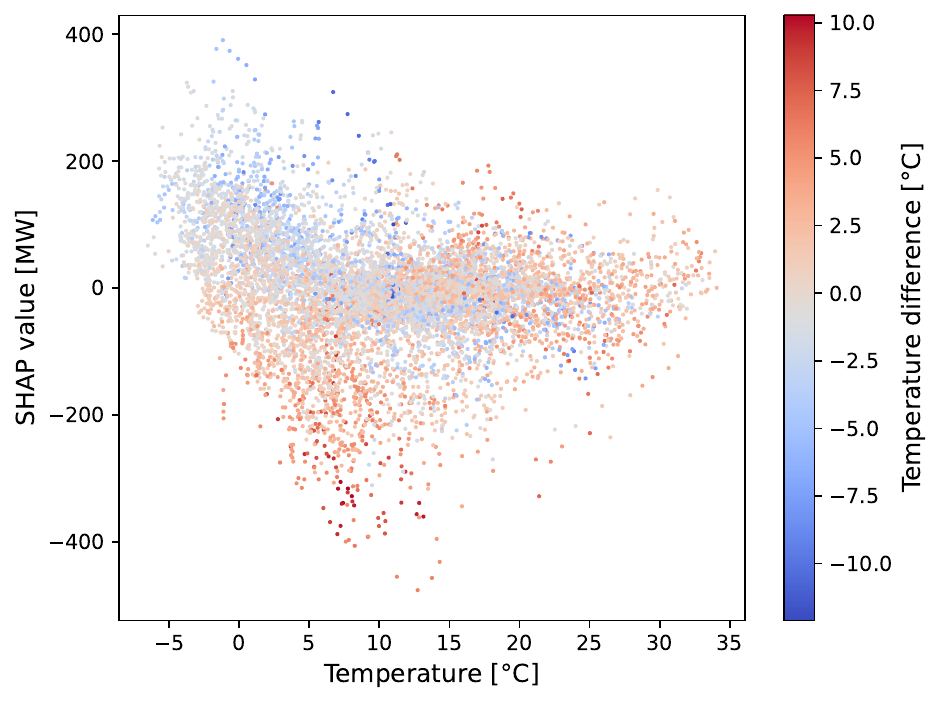}
        \label{fig:temperature-tabpfn}
    \end{subfigure}\\ \vspace{-0.5cm}
    \begin{subfigure}{0.49\textwidth}
        \includegraphics[width=\textwidth]{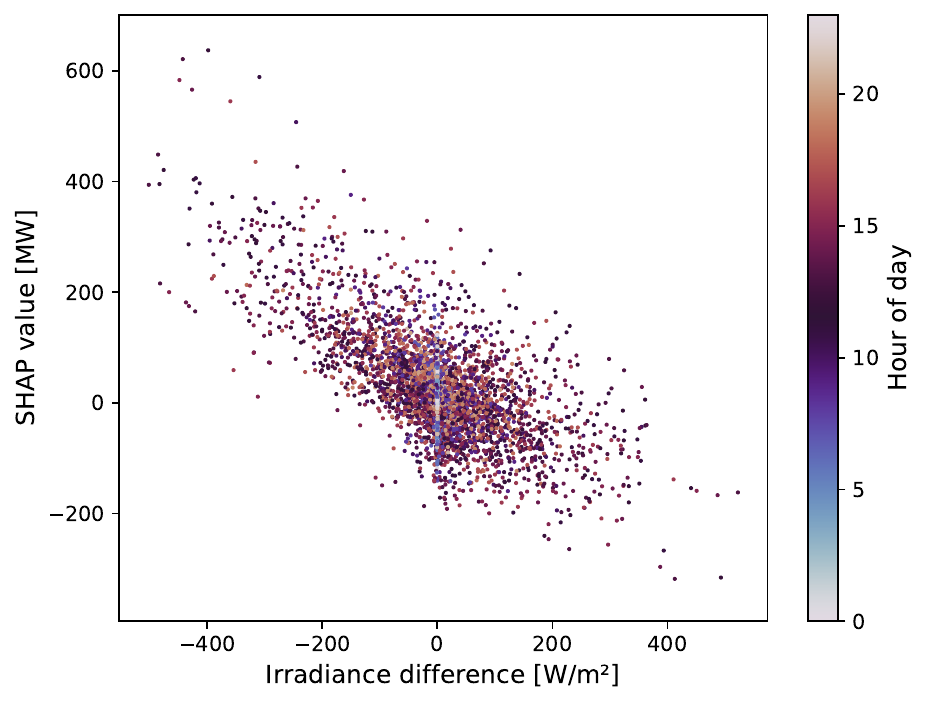}
        \label{fig:irradiance-chronos2}
    \end{subfigure}
    \begin{subfigure}{0.49\textwidth}
        \includegraphics[width=\textwidth]{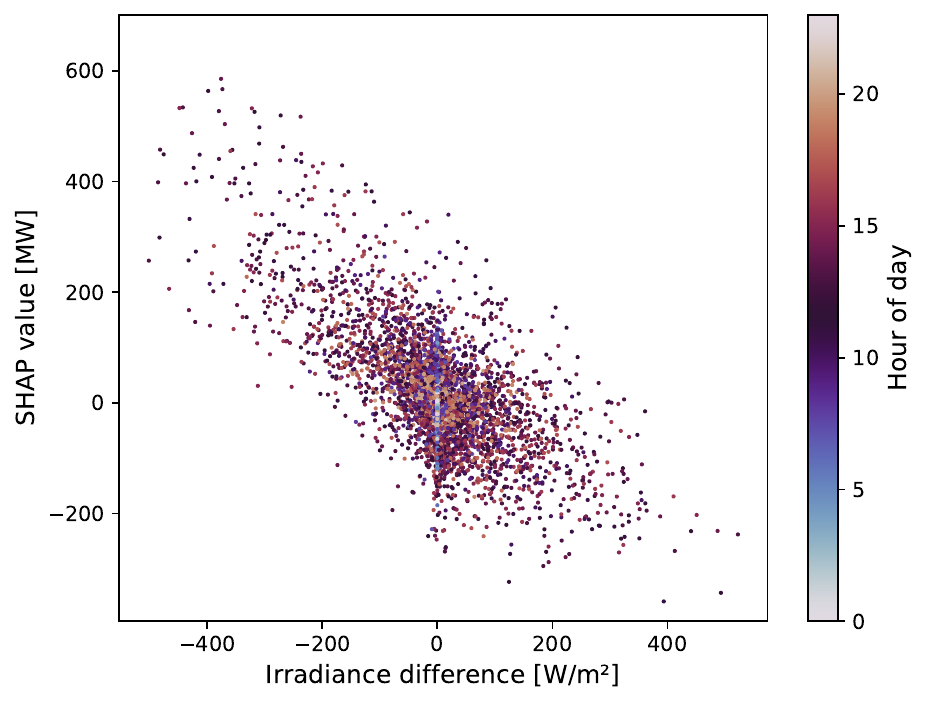}
        \label{fig:irradiance-tabpfn}
    \end{subfigure}
    \vspace{-0.5cm}
    \caption{Dependence of Chronos-2 (left) and TabPFN-TS (right) on the holiday covariate (top), the temperature covariate (middle) and the irradiance difference to the previous day (bottom). Both TSFMs make similar use of the covariates and this use is consistent with domain knowledge. The predicted load is decreased on holidays (top), increased when the outside temperature drops on cold days (middle), and increased when PV generation drops due to lower irradiance (bottom).}
    \Description{Dependence of Chronos-2 (left) and TabPFN-TS (right) on the holiday covariate (top), the temperature covariate (middle) and the irradiance difference to the previous day (bottom). Both TSFMs make similar use of the covariates and this use is consistent with domain knowledge. The predicted load is decreased on holidays (top), increased when the outside temperature drops on cold days (middle), and increased when PV generation drops due to lower irradiance (bottom).}
    \label{fig:feature-dependence}
\end{figure*}

\begin{itemize}
    \item \emph{Holiday dependence}:
    The SHAP values are shown for six categories of days: Sunday, Monday, a workday following a workday ($\circ/\circ$), a workday following a holiday ($\bullet/\circ$), a holiday following a workday ($\circ/\bullet$), and a holiday following a holiday ($\bullet/\bullet$).
    When the previous day was a holiday and the next day is no holiday ($\bullet/\circ$), the holiday feature increases the predicted load. When the next day is a holiday ($\circ/\bullet$ and $\bullet/\bullet$), it decreases the load instead.
    The effect is stronger during the day (dark dots) than during the night (bright dots), where the base load is similar on all weekdays.
    This aligns well with our expectation that the load is reduced on holidays due to less industrial and economic activity, and jumps back to a normal level after the holiday.
    Chronos-2 is only slightly affected by the holiday feature on Sundays and Mondays, whereas TabPFN-TS shows a pronounced influence of the holiday feature on these days. This signifies that TabPFN-TS uses the holiday feature to predict a lower load on Sundays (where the feature is set to 1) and a higher load again on Mondays, whereas Chronos-2 infers the weekly pattern from the past load.
    \item \emph{Temperature dependence}: For temperatures below 15\,°C, both models react to temperature differences with respect to the previous day.
    When the temperature increases with respect to the temperature 24 hours ago (red dots), the predicted load decreases, resulting in negative SHAP values. When the temperature decreases (blue dots), the opposite effect is observed, resulting in positive SHAP values.
    This can be plausibly explained by the increased use of electric heating on cold days.
    At warmer temperatures, the magnitude of the SHAP values is smaller and the effect of the temperature difference is less pronounced.
    This temperature dependence is further illustrated by visualizing SHAP values against both current and previous day temperatures, confirming that both TSFMs rely more heavily on this covariate during periods of pronounced change, as shown in Figure~\ref{fig:dependence-temperature-difference} of Appendix~\ref{app:global-explanations}.
    \item \emph{Irradiance dependence}: The lower panels show the SHAP values depending on the difference in the irradiance from the day before. 
    Unsurprisingly, the difference is zero at night (bright dots), because the sun does not shine. 
    During the day (dark dots), an almost linear pattern is visible for both models, with the slope directly quantifying how much the predicted load changes per unit change in irradiance.
    The models use the irradiance feature to react to changing weather conditions -- when the irradiance decreases (negative $\Delta$irradiance), the predicted load increases (positive SHAP value), and vice versa.
    This can be explained by the fact that the TSO load is affected by behind-the-meter photovoltaic, so that a higher irradiance decreases the effective load.
\end{itemize}
Overall, the covariate dependence of the TSFMs aligns well with our expectations derived from domain knowledge about influences of holidays, temperature and irradiance on the TSO load, indicating that the TSFMs make meaningful use of the covariates.

\subsection{Local explanations}

Next, we examine the local explanations for both past load windows and covariates to give further insights into when particular covariates are used by the TSFMs.
We highlight four representative examples for the Chronos-2 model in Figure~\ref{fig:local-explanations} that illustrate the models' behavior across different scenarios.
The complete set of local explanations for both models for the entire test period is provided in Appendix~\ref{app:local-explanations} (Figure~\ref{fig:monthly-shap-values-chronos} for Chronos-2 and Figure~\ref{fig:monthly-shap-values-tabpfn} for TabPFN-TS). 
Since Chronos-2 and TabPFN-TS exhibit similar patterns, Figure~\ref{fig:local-explanations} shows the effects only based on Chronos-2 (see Figure~\ref{fig:local-explanations-tabpfn} in the appendix for the same examples with TabPFN-TS):

\begin{itemize}
    \item \emph{Post-holiday effects} (Figure~\ref{fig:local-explanations} (a)):
    On January 6, \ifanonymize{which is a public holiday}{a public holiday in Baden-Württemberg}, the holiday feature shows a negative SHAP value, indicating that the model appropriately reduces the predicted load. 
    The following day (January 7) marks the first workday after an extended holiday period. 
    While the holiday feature still plays a role by contributing positively to increase predicted load (reflecting the return to workdays), the previous day has a negative influence, resulting in an underestimation of the load on this transition day.
    By January 8, the model generates more accurate predictions by using January 7 as a reference -- indicated by the large influence of the last day on January 8. 
    This shows how the model uses the immediate post-holiday day as a marker for the end of the Christmas period.
    
     \item \emph{Temperature effects} (Figure~\ref{fig:local-explanations} (b)): 
     Rising temperatures on February 20-21 produce negative SHAP values for the temperature feature, appropriately reducing the predicted load, which aligns with the inverse temperature-load relationship we would expect from domain knowledge that a part of the heating load is electric.
     
     \item \emph{Holiday effects} (Figure~\ref{fig:local-explanations} (c)): 
     The May 1 example illustrates the impact of holidays. 
     The holiday feature decreases the load on May 1 (a Thursday) and increases it on May 2 to account for the return to normal working patterns the following day. 
     Similar to January 7, the previous day negatively impacts the prediction. 
     This shows that the holiday feature helps the model to predict that the low load on the previous day is not due to a permanent change in consumption levels, but the load will increase again on the next workday.
    
    \item \emph{Irradiance effects} (Figure~\ref{fig:local-explanations} (d)): 
    On August 20, a lower solar irradiance compared to the previous day leads to adjustments in the midday load curve. This indicates that the model captures the dependency between solar generation and load. Reduced behind-the-meter photovoltaic output lowers self-consumption and increases the net load, which is reflected in a higher prediction.
\end{itemize}

\begin{figure*}[hbt]
    \centering
    \includegraphics[width=\linewidth]{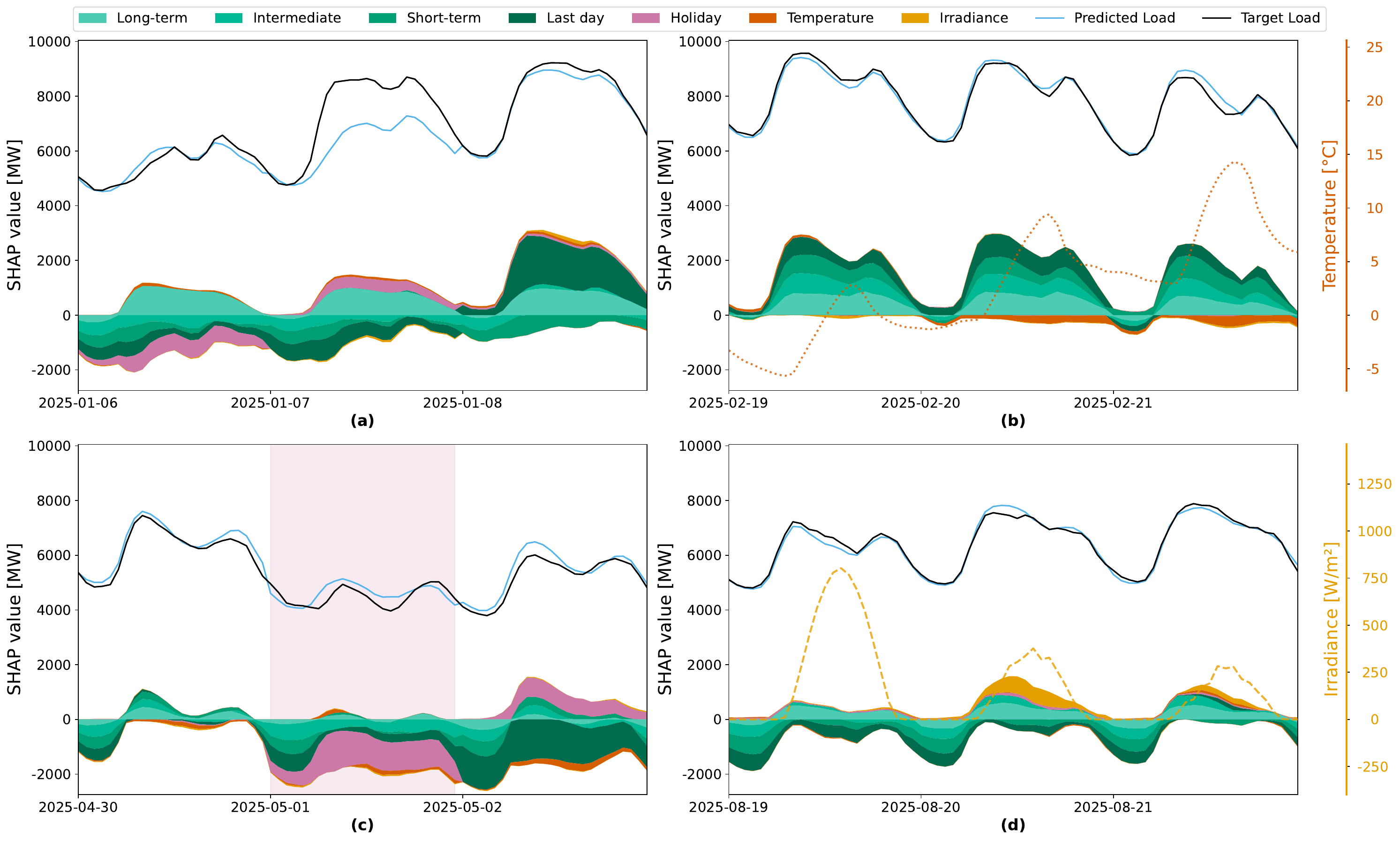}
    \caption{Local SHAP explanations for Chronos-2 predictions on selected days showing feature contributions to load forecasts: (a) post-holiday effects (January 6-8), (b) temperature responses (February 20-21), (c) holiday patterns (May 1-2), and (d) irradiance influence (August 20). Positive (negative) SHAP values correspond to an increase (decrease) in the predicted load.}
    \Description{SHAP explanation panel for Chronos-2 predictions on four selected days: (a) post-holiday effects (January 6-8), (b) temperature responses (February 20-21), (c) holiday patterns (May 1-2), and (d) irradiance influence (August 20). Positive SHAP values indicate increased predicted load, negative values indicate decreased predicted load.}
    \label{fig:local-explanations}
\end{figure*}


An analysis of the complete test period (Appendix~\ref{app:local-explanations}, Figures~\ref{fig:monthly-shap-values-chronos} and~\ref{fig:monthly-shap-values-tabpfn}) reveals several broader patterns in the importance of covariates. Covariates contribute most strongly on specific days rather than uniformly over time. In particular, their importance increases when temperature or irradiance differs substantially from the previous day, or when the preceding or following day is a holiday. Furthermore, the previous day’s load is especially influential for forecasts on Tuesdays, while its importance is markedly lower on Mondays and weekends. This pattern indicates that the model captures weekly cyclical patterns in electricity consumption behavior, distinguishing between workdays and weekends.

\section{Conclusion \& Future Work}
\label{sec:conclusion}

The present work demonstrates that Time Series Foundation Models (TSFMs) offer a promising combination of accuracy and efficiency for load forecasting applications, and that they can be meaningfully explained when paired with efficient XAI approaches such as the one proposed here.
We show that Chronos-2 and TabPFN-TS achieve competitive performance with Transformers trained from scratch on domain-specific data, while requiring no training data or computational resources for model development — making accurate forecasting accessible even in scenarios where training data is scarce.
The flexibility of TSFMs to operate with varying context lengths and covariates enables efficient computation of exact SHAP values, with full explanations generated in approximately 5 seconds for Chronos-2.
Our global and local explanation analysis reveals that TSFMs utilize covariates in ways that align with established domain knowledge of energy consumption patterns, including temperature dependencies, calendar effects, and post-holiday phenomena.
Our work demonstrates that TSFMs can satisfy both the accuracy and explainability demands of critical infrastructure applications while preserving their zero-shot advantages.

We identify several promising research directions at the intersection of XAI with TSFMs. 
Applying our methodology to other energy forecasting tasks -- including low-voltage load, price, and renewable generation forecasting -- would demonstrate the generalizability of our approach across diverse operational contexts.
Since our approach essentially only requires models to accept masked inputs, a capability inherent to most TSFM architectures, it is largely model-agnostic and can be easily generalized to other foundation models. 
Applying our SHAP adaptation to other TSFMs, both univariate (e.g.,~\cite{liu_aksu_2026}) and covariate-informed (e.g.,~\cite{cohen_khwaja_2025}), would therefore provide broader insights into how different architectural choices affect feature attribution patterns in TSFMs.
Computational efficiency could be substantially improved by estimating SHAP values from subsets of feature group combinations rather than exhaustively evaluating all $2^N$ coalitions -- enabling finer-grained temporal groupings beyond the current four buckets, which is a particularly important consideration for the slower TabPFN-TS and scenarios with larger feature sets.
Extending our framework to explain probabilistic forecasts and uncertainty quantification would address a critical gap for risk-sensitive operational decisions~\cite{nikoltchovska_putz_2025}.
Finally, analyzing how fine-tuning affects feature attribution patterns could reveal what domain-specific knowledge is acquired during adaptation and inform best practices for model deployment in specialized forecasting contexts.

\section*{Data \& Code Availability}

The dataset and the code to reproduce the forecasts and explanations for Chronos-2 and TabPFN-TS are available at \url{https://github.com/KIT-IAI/SHAP4TSFMs}. The original TSFM implementations and weights are available with the publications~\cite{ansari_chronos-2_2025, hoo_tables_2025}.

\section*{Statement on the Usage of Generative AI}

ChatGPT with GPT-5.2 and Claude Sonnet 4.5 were used to improve the fluency of the text. The authors have checked all produced text carefully and take the full responsibility for the submitted manuscript.

\begin{acks}
The authors gratefully acknowledge funding from the Helmholtz Association under the program “Energy System Design” and the Networking Fund through Helmholtz AI and the HAICORE@KIT partition. 
We thank Jonathan Kolar for the data preprocessing and Alexander Kreusel for helpful discussions about TSFMs.
\end{acks}

\bibliographystyle{ACM-Reference-Format}
\bibliography{zotero-references-an, zotero-references}

\appendix
\section{Extended Methodology}
\label{app:methods}
\paragraph{Chronos-2} 
Chronos-2~\cite{ansari_chronos-2_2025} is a pretrained foundation model for time series forecasting capable of handling univariate, multivariate, and covariate-informed forecasting tasks. 
The model architecture tokenizes time series data, enabling it to process temporal sequences through a transformer-based framework. 
For covariate-informed forecasting, Chronos-2 builds patches for each variable and uses group attention to incorporate information from the covariates.
This design allows the model to learn cross-channel dependencies while leveraging its pretrained representations for zero-shot and few-shot forecasting scenarios.

\paragraph{TabPFN-TS} 
TabPFN-TS~\cite{hoo_tables_2025} extends TabPFN~\cite{hollmann_tabpfn_2023}, a foundation model originally designed for classification and regression for tabular data, to the time series forecasting domain. 
Rather than treating forecasting as a sequential modeling problem, TabPFN-TS reframes it as a tabular supervised learning task where historical observations and future time steps are represented in a tabular format. 
This approach is particularly effective for small-to-medium sized datasets common in energy forecasting applications. 
The model explicitly handles covariates as separate input features within the tabular representation.
This enables direct integration of weather variables, calendar information, and other contextual data alongside historical load patterns. 

\section{Extended Results}
\subsection{Evaluation Metrics}
\label{app:metrics}
We evaluate forecasting performance using three standard regression metrics. Let $y_i$ denote the actual load at time step $i$, $\hat{y}_i$ the predicted load, and $n$ the number of predictions.

Mean absolute error (MAE) measures the average absolute deviation between predictions and actual values:
\begin{equation}
\text{MAE} = \frac{1}{n}\sum_{i=1}^{n}|y_i - \hat{y}_i|
\end{equation}

Root mean squared error (RMSE) penalizes larger errors more heavily due to the quadratic term:
\begin{equation}
\text{RMSE} = \sqrt{\frac{1}{n}\sum_{i=1}^{n}(y_i - \hat{y}_i)^2}
\end{equation}

Mean absolute percentage error (MAPE) expresses errors as percentages relative to actual values:
\begin{equation}
\text{MAPE} = \frac{100\%}{n}\sum_{i=1}^{n}\left|\frac{y_i - \hat{y}_i}{y_i}\right|.
\end{equation}

\subsection{Forecasting Performance with Different Context Lengths}
\label{app:context-lengths}

Figure~\ref{fig:performance} illustrates how forecasting performance varies with context length for both TSFMs. 
TabPFN-TS shows substantial improvement as context increases, with MAE decreasing from 549~MW at one week to 163.2~MW at 52 weeks of context. 
In contrast, Chronos-2 achieves near-optimal performance already at 16 weeks of context, plateauing around 156~MW MAE. 
Both models surpass the persistence baseline (325.9~MW MAE) across all context lengths, with Chronos-2 demonstrating superior performance and greater sample efficiency in utilizing historical information.

\begin{figure}[ht]
    \centering
    \includegraphics[width=\linewidth]{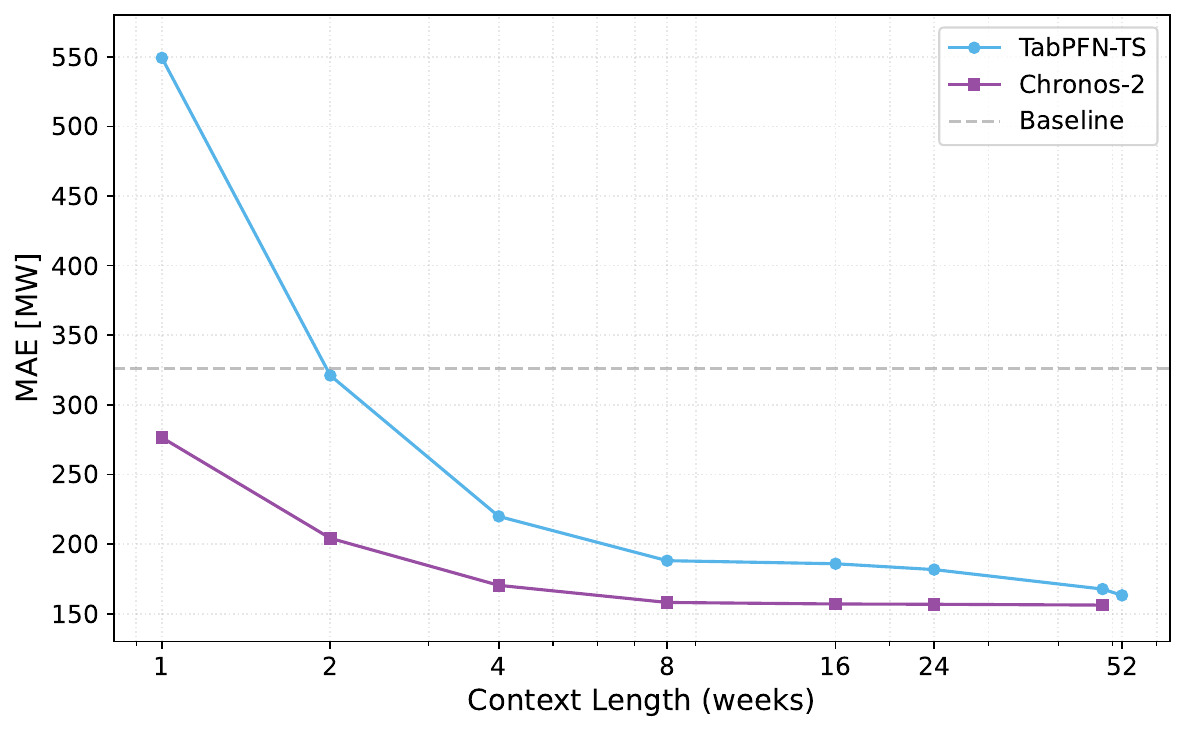}
    \caption{Performance comparison of the TabPFN-TS and Chronos-2 models across different context lengths.}
    \Description[Performance comparison of Chronos-2 and TabPFN-TS with different context lengths.]{Line plot comparing MAE performance of the Chronos-2 and TabPFN-TS foundation models across context lengths from 1 to 52 weeks on a logarithmic x-axis. TabPFN-TS (light blue circles) starts at 515 MW MAE with 1 week context and decreases to 169.5 MW at 52 weeks. Chronos-2 (purple squares) starts at 277 MW with 1 week context and plateaus around 156 MW after 16 weeks. A horizontal gray dashed line at 326 MW represents the MAE of the baseline persistence forecast.}    
    \label{fig:performance}
\end{figure}

\subsection{Global Explanations - Dependence on Temperature Difference}
\label{app:global-explanations}

Figure~\ref{fig:dependence-temperature-difference} presents the SHAP values for the temperature covariate depending on the temperature of the predicted day and the temperature of the day before. The black line indicates that the temperature is equal to the day before. Points above the line indicate a temperature increase, and points below the line indicate a temperature decrease. Chronos-2 decreases the predicted load with increasing temperatures, shown by the negative SHAP values above the line. TabPFN-TS does so only for temperatures below 15\,°C.

\begin{figure*}
    \centering
    \begin{subfigure}{0.49\textwidth}
        \caption{Chronos-2}
        \includegraphics[width=\textwidth]{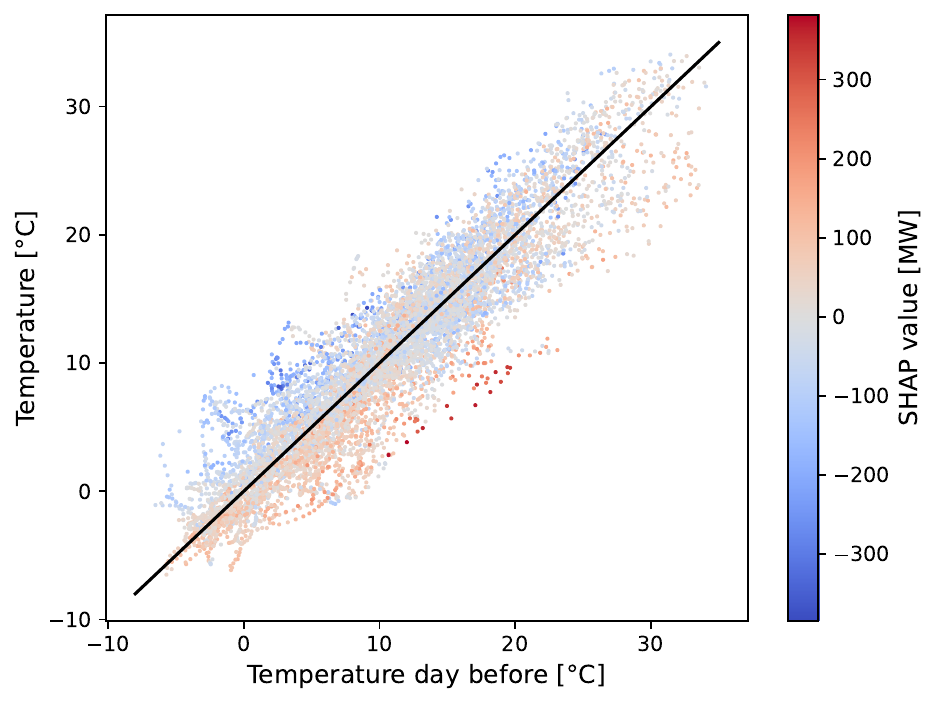}
        \label{fig:temperature-day-before-chronos2}
    \end{subfigure}
    \begin{subfigure}{0.49\textwidth}
        \caption{TabPFN-TS}
        \includegraphics[width=\textwidth]{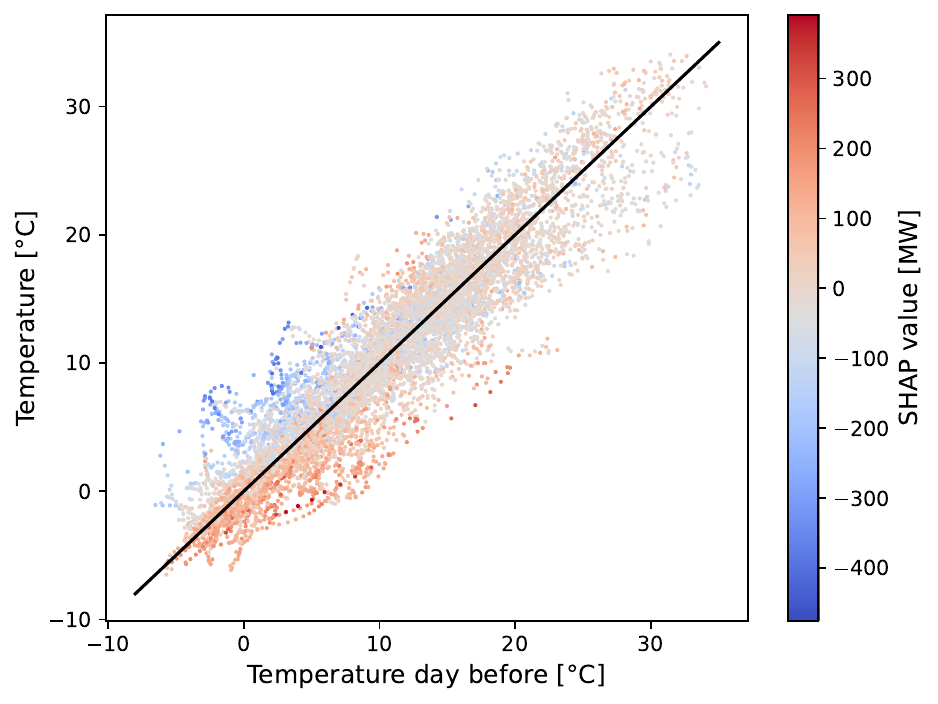}
        \label{fig:temperature-day-before-tabpfn}
    \end{subfigure}
    \caption{SHAP dependence plots of temperature difference between prediction day and previous day. The back line indicates points where the temperature is equal to the previous day. Both models decrease the predicted load with increasing temperatures (points above the black line), and vice versa (points below the black line).}
    \Description{SHAP dependence plots of temperature difference between prediction day and previous day. The back line indicates points where the temperature is equal to the previous day. Both models decrease the predicted load with increasing temperatures (points above the black line), and vice versa (points below the black line).}
    \label{fig:dependence-temperature-difference}
\end{figure*}

\subsection{Local Explanations - TabPFN-TS Examples}
\label{app:local-explanations}

Figure~\ref{fig:local-explanations-tabpfn} provides detailed local explanations for TabPFN-TS across representative forecasting scenarios. 
These examples demonstrate how feature contributions vary with specific operational contexts: post-holiday load recovery (January 6--8), temperature-driven demand variations (February 20--21), holiday effects (May 1--2), and solar irradiance influence on summer demand (August 20). 
The explanations reveal consistent patterns aligned with energy domain knowledge, including the model's ability to distinguish between different types of calendar effects and its appropriate weighting of weather variables under varying seasonal conditions.

\begin{figure*}
    \centering
    \includegraphics[width=\linewidth]{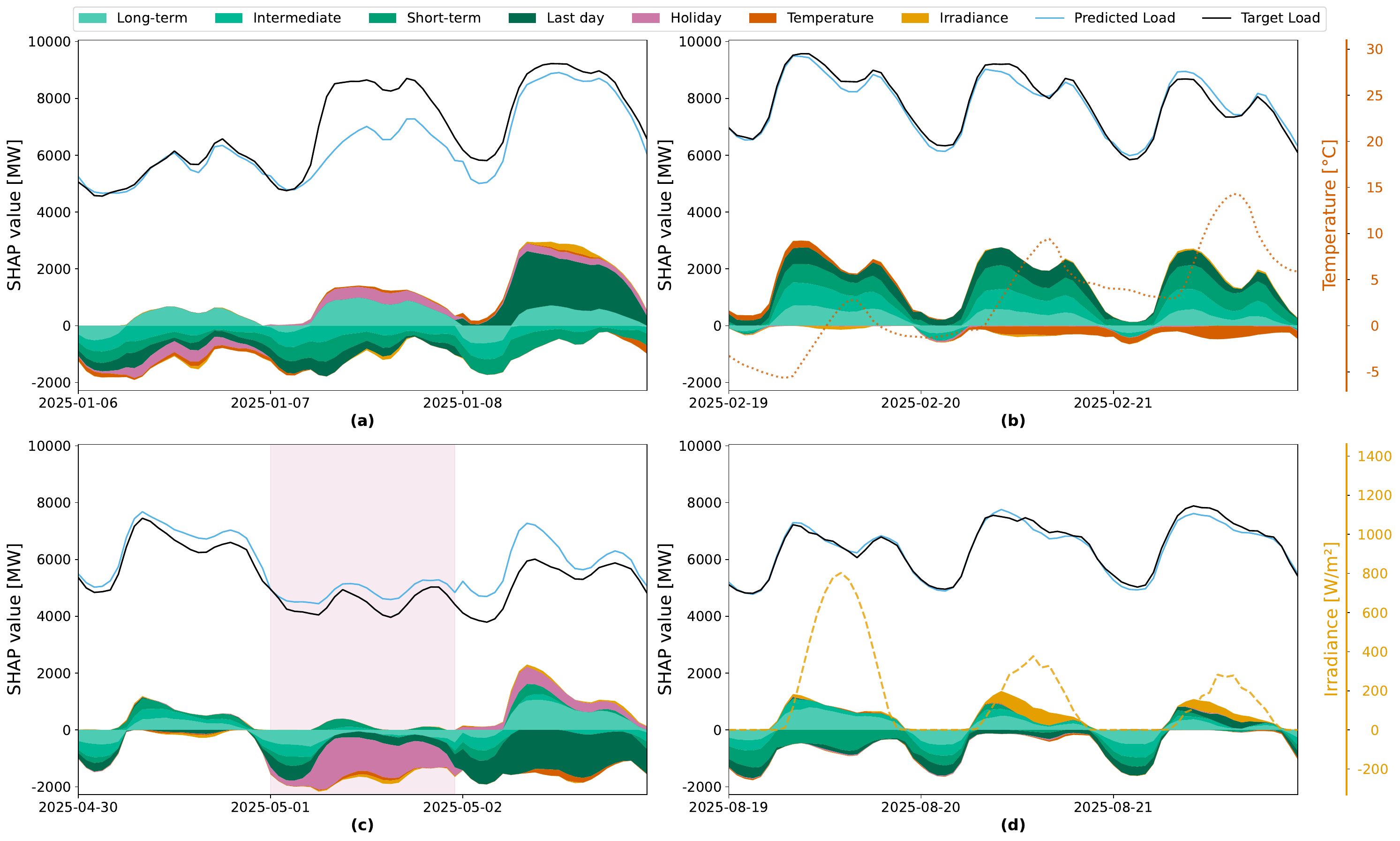}
    \caption{Local SHAP explanations for TabPFN-TS predictions on selected days showing feature contributions to load forecasts: (a) post-holiday effects (January 6-8), (b) temperature responses (February 20-21), (c) holiday patterns (May 1-2), and (d) irradiance influence (August 20). Positive (negative) SHAP values correspond to an increase (decrease) in the predicted load.}
    \Description{Local SHAP explanations for TabPFN-TS predictions on selected days showing feature contributions to load forecasts: (a) post-holiday effects (January 6-8), (b) temperature responses (February 20-21), (c) holiday patterns (May 1-2), and (d) irradiance influence (August 20). Positive (negative) SHAP values correspond to an increase (decrease) in the predicted load.}
    \label{fig:local-explanations-tabpfn}
\end{figure*}

\subsection{Local Explanations - SHAP values of one year}

Figures~\ref{fig:monthly-shap-values-chronos} and~\ref{fig:monthly-shap-values-tabpfn} present local SHAP explanations across the complete test set for Chronos-2 and TabPFN-TS, respectively and complement the focused examples in Section~\ref{sec:explanations}.
Each figure displays twelve monthly panels (October 2024--September 2025) showing model predictions, SHAP feature attributions, and actual covariate values.

The visualizations reveal how feature importance varies across seasonal cycles, major holidays, and weather events throughout the year. Temperature attributions show strong seasonal patterns, with negative contributions during summer months (reduced heating demand) and positive contributions in winter. Irradiance attributions peak during summer, reflecting increased cooling loads on sunny days. The holiday covariate captures substantial load reductions during public holidays, especially around major holidays such as New Year's Day, Easter, and Christmas.

\begin{figure*}
    \centering
    \includegraphics[width=0.94\textwidth]{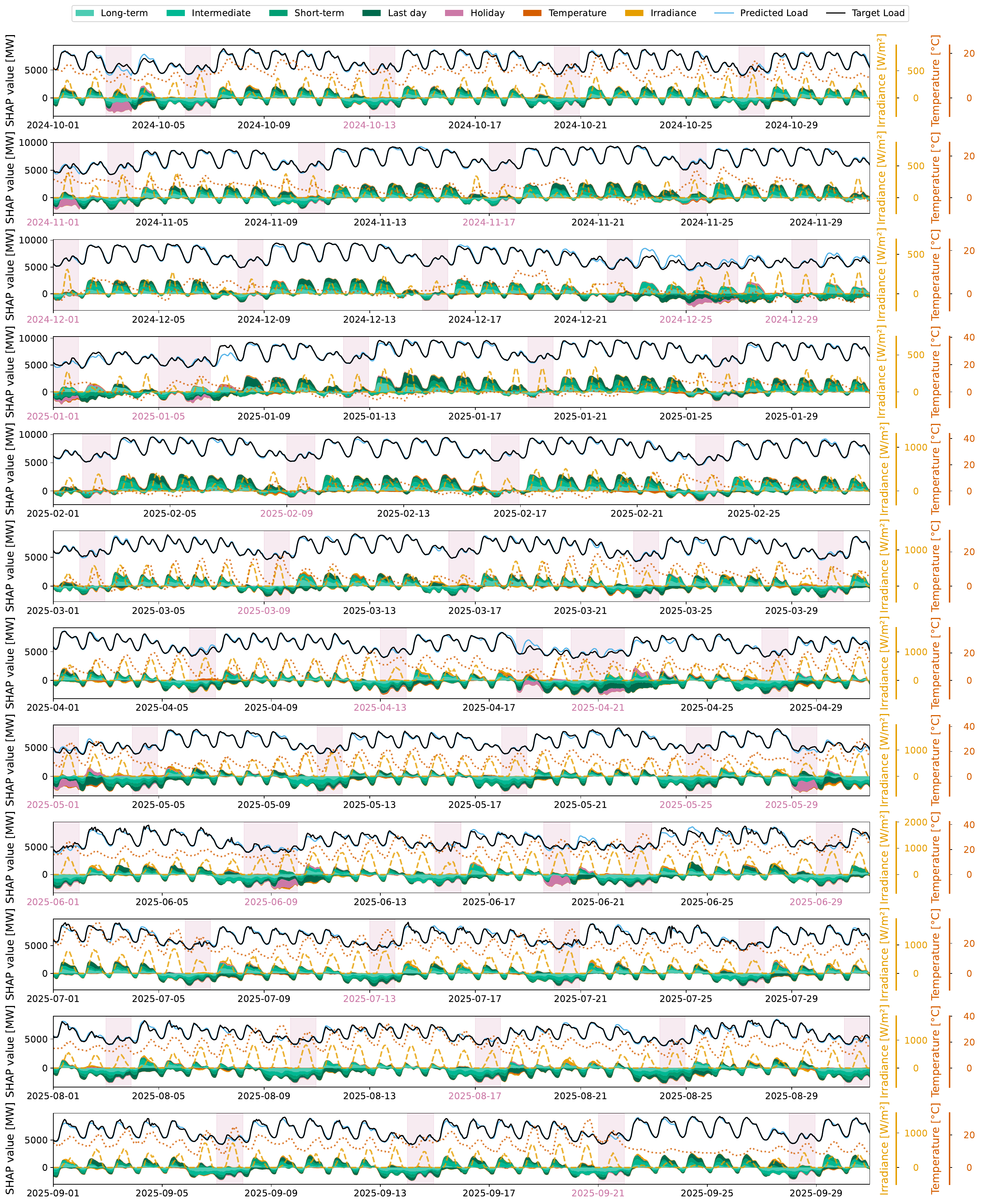}
    \vspace{-0.3cm}
    \caption{Chronos-2 predictions and local SHAP feature attributions across the full test set, with one panel per month. Each panel shows: actual load (black), model predictions (blue); SHAP values indicating feature contributions to load forecasts, where positive (negative) values correspond to increased (decreased) predicted load; actual covariate values for temperature and irradiance (right y-axis), with Sundays and holidays marked in pink.}
    \Description{Chronos-2 predictions and local SHAP feature attributions across the full test set, with one panel per month. Each panel shows: actual load (black), model predictions (blue); SHAP values indicating feature contributions to load forecasts, where positive (negative) values correspond to increased (decreased) predicted load; actual covariate values for temperature and irradiance (right y-axis), with Sundays and holidays marked in pink.}
    \label{fig:monthly-shap-values-chronos}
\end{figure*}

\begin{figure*}
    \centering
    \includegraphics[width=0.94\textwidth]{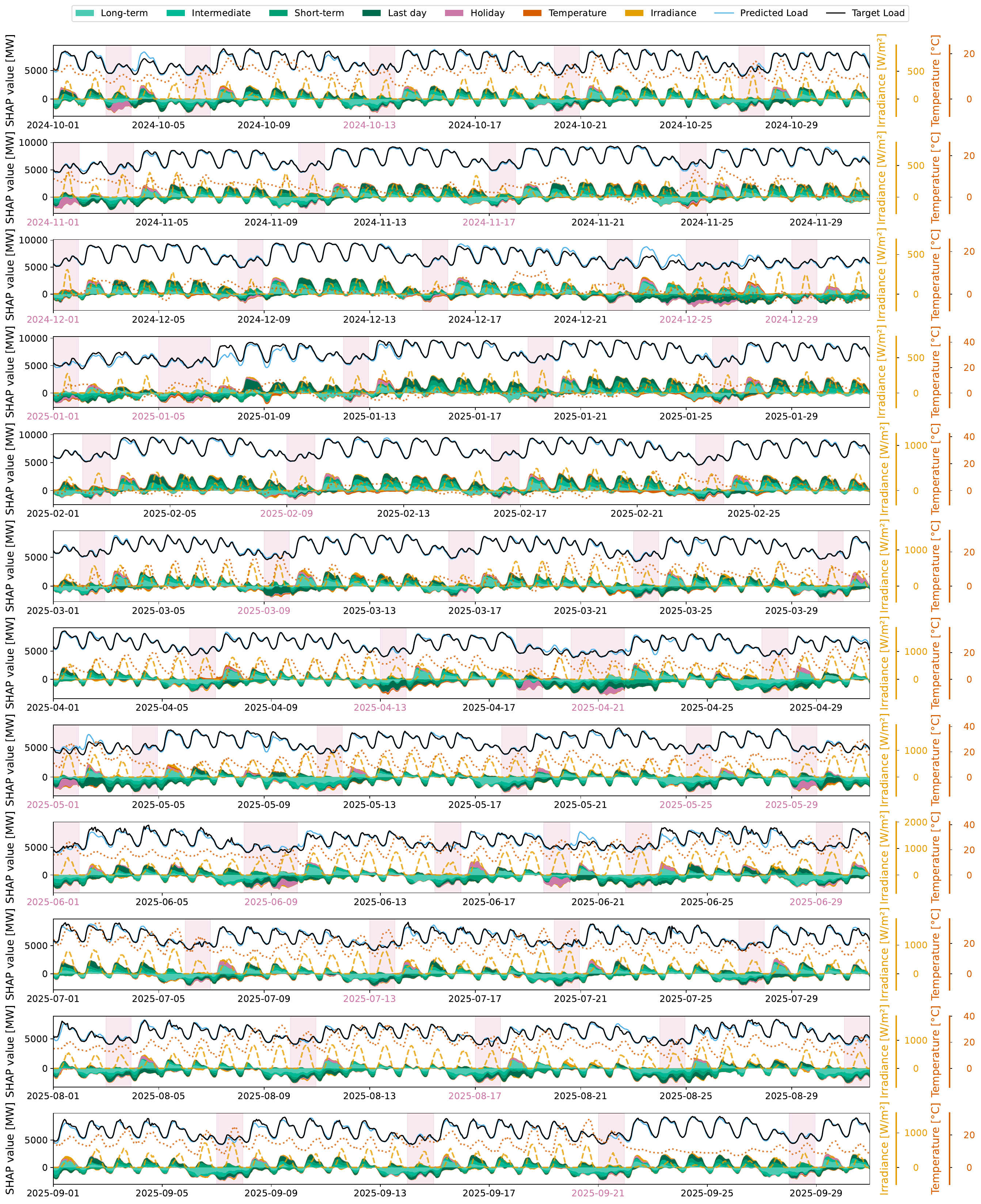}
    \vspace{-0.3cm}
    \caption{TabPFN-TS predictions and local SHAP feature attributions across the full test set, with one panel per month. Each panel shows: actual load (black), model predictions (blue); SHAP values indicating feature contributions to load forecasts, where positive (negative) values correspond to increased (decreased) predicted load; actual covariate values for temperature and irradiance (right y-axis), with Sundays and holidays marked in pink.}
    \Description{TabPFN-TS predictions and local SHAP feature attributions across the full test set, with one panel per month. Each panel shows: actual load (black), model predictions (blue); SHAP values indicating feature contributions to load forecasts, where positive (negative) values correspond to increased (decreased) predicted load; actual covariate values for temperature and irradiance (right y-axis), with Sundays and holidays marked in pink.}
    \label{fig:monthly-shap-values-tabpfn}
\end{figure*}

\end{document}
\endinput